\documentclass[letterpaper,10pt,conference]{ieeeconf}
\IEEEoverridecommandlockouts
\usepackage[T1]{fontenc}

\usepackage{graphicx}
\makeatletter
\newcommand{\tablebodyfont}{\footnotesize}
\newcommand{\sunrgbdimagefull}{\includegraphics[width=0.88\textwidth]{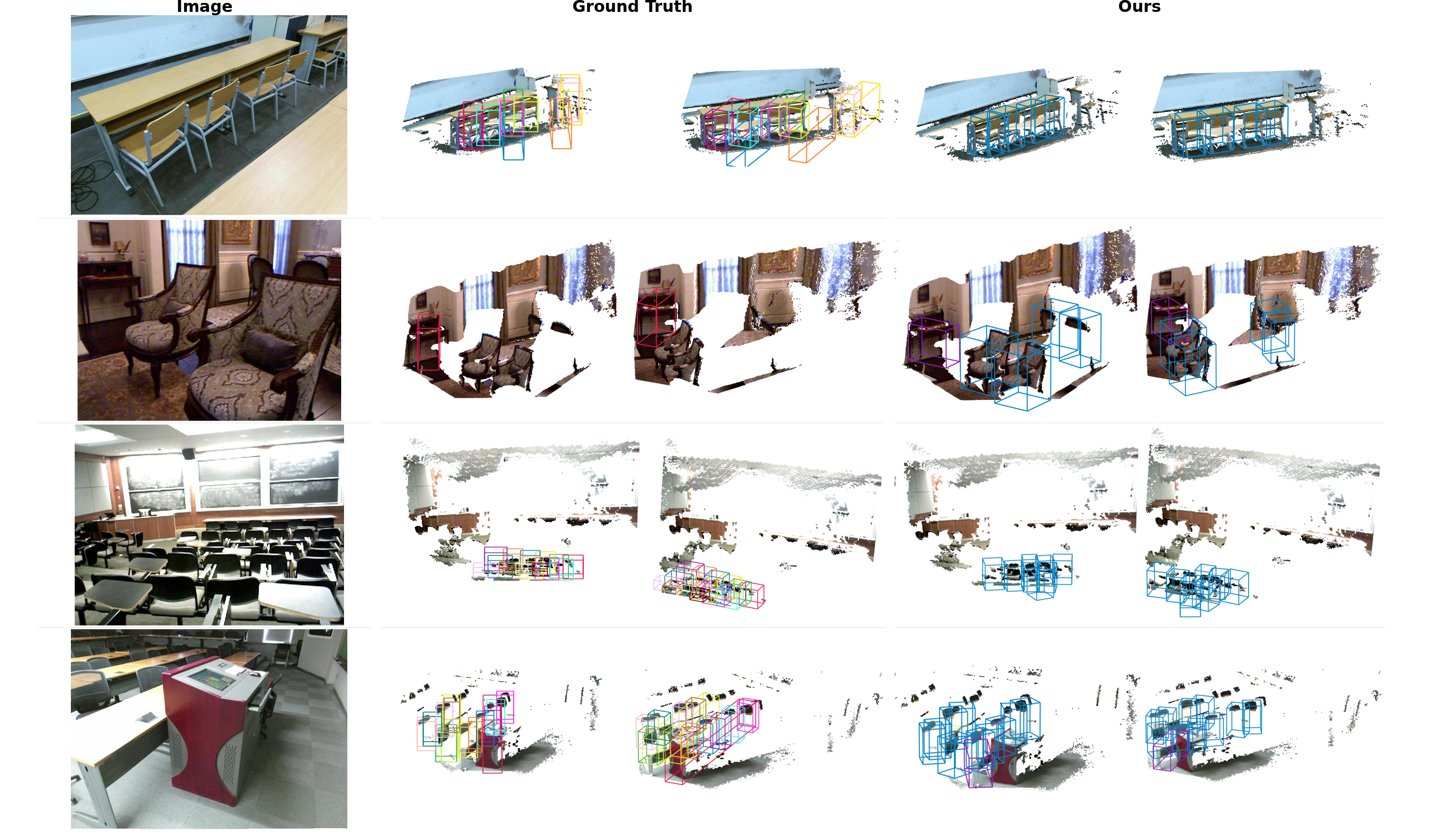}}
\newcommand{\scannetimagefull}{\includegraphics[width=0.88\textwidth]{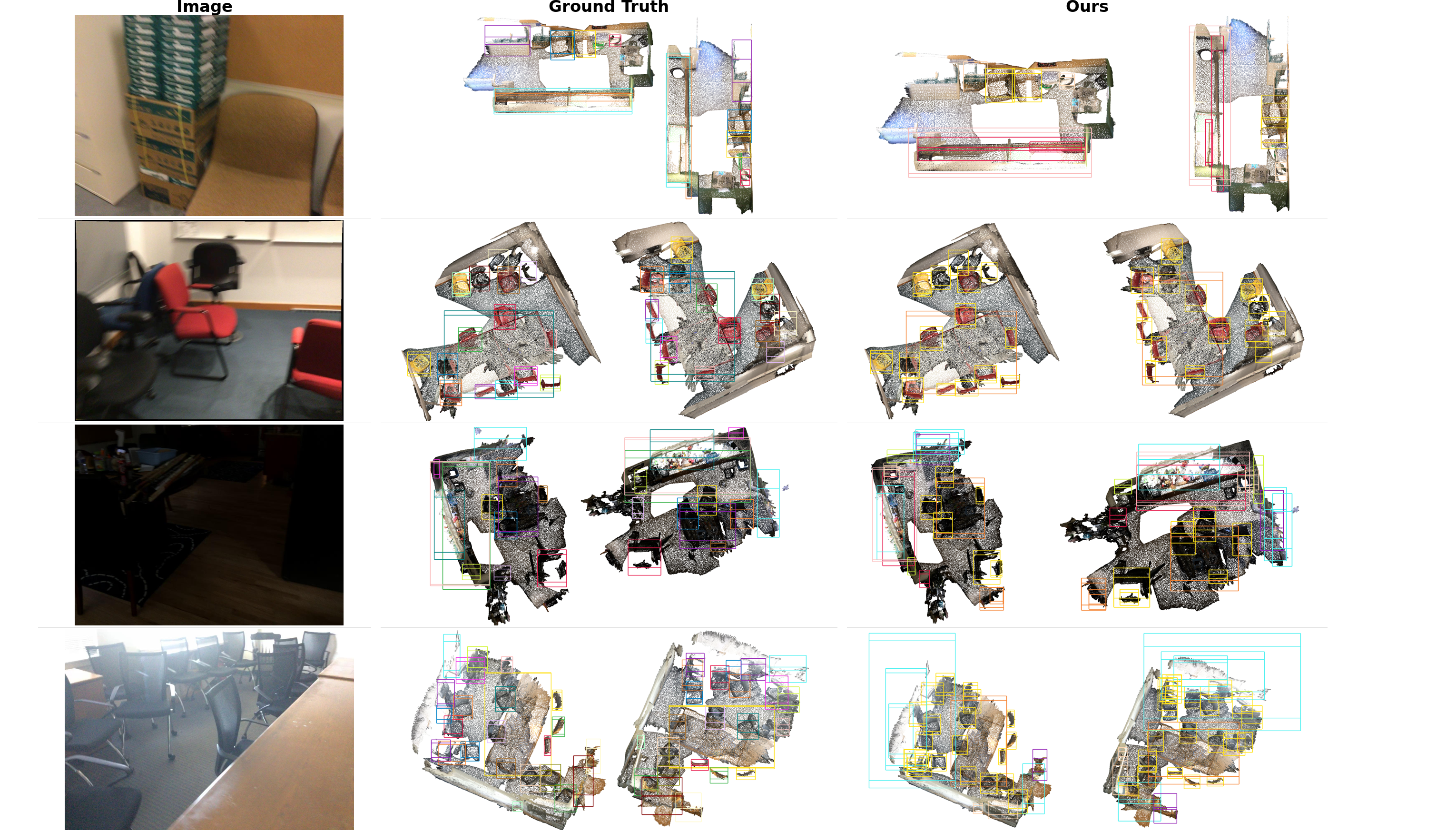}}
\newcommand{\pipelineimagefull}{\includegraphics[width=0.86\textwidth]{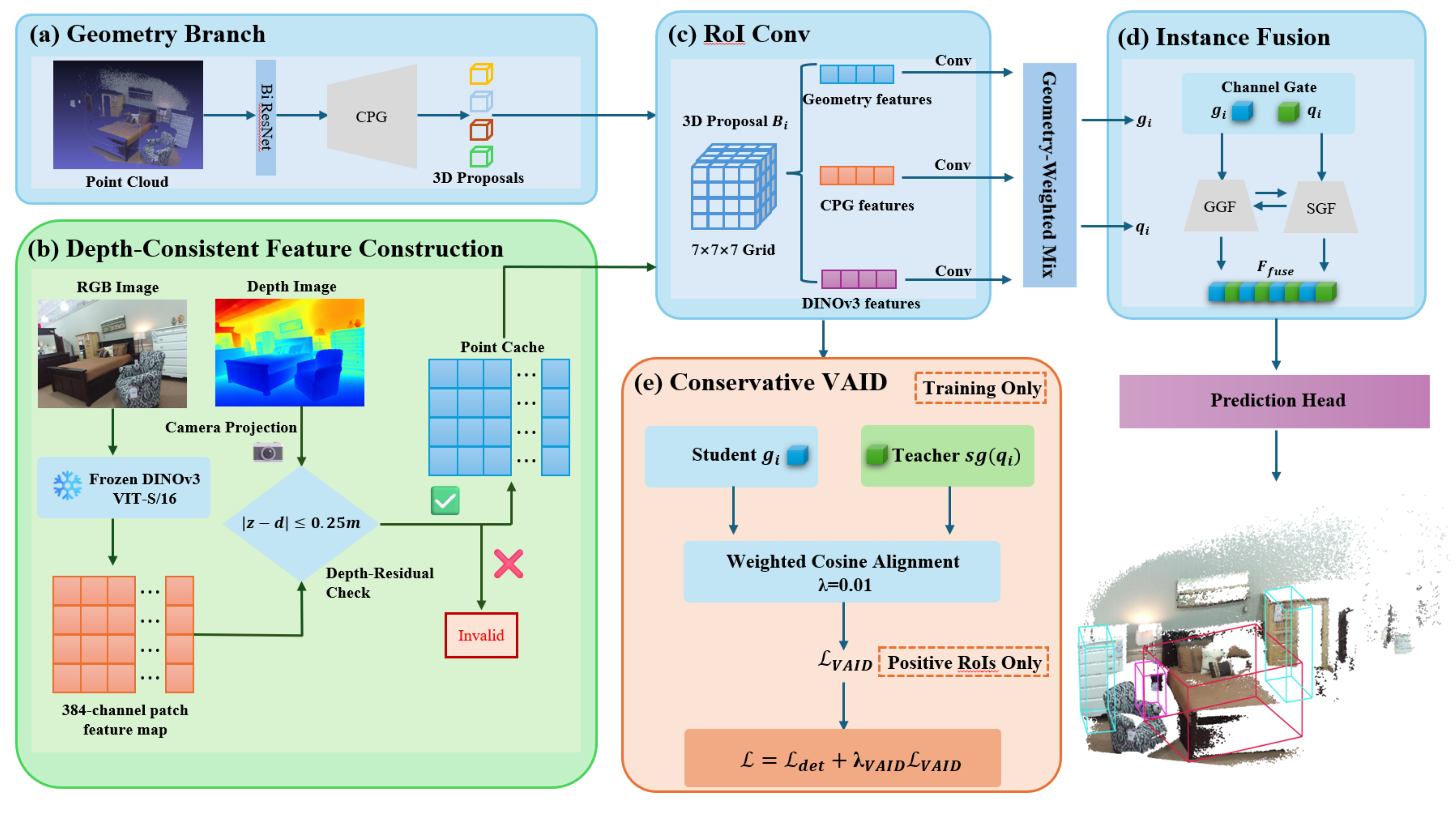}}
\newcommand{\depthimagefull}{\includegraphics[width=0.92\textwidth]{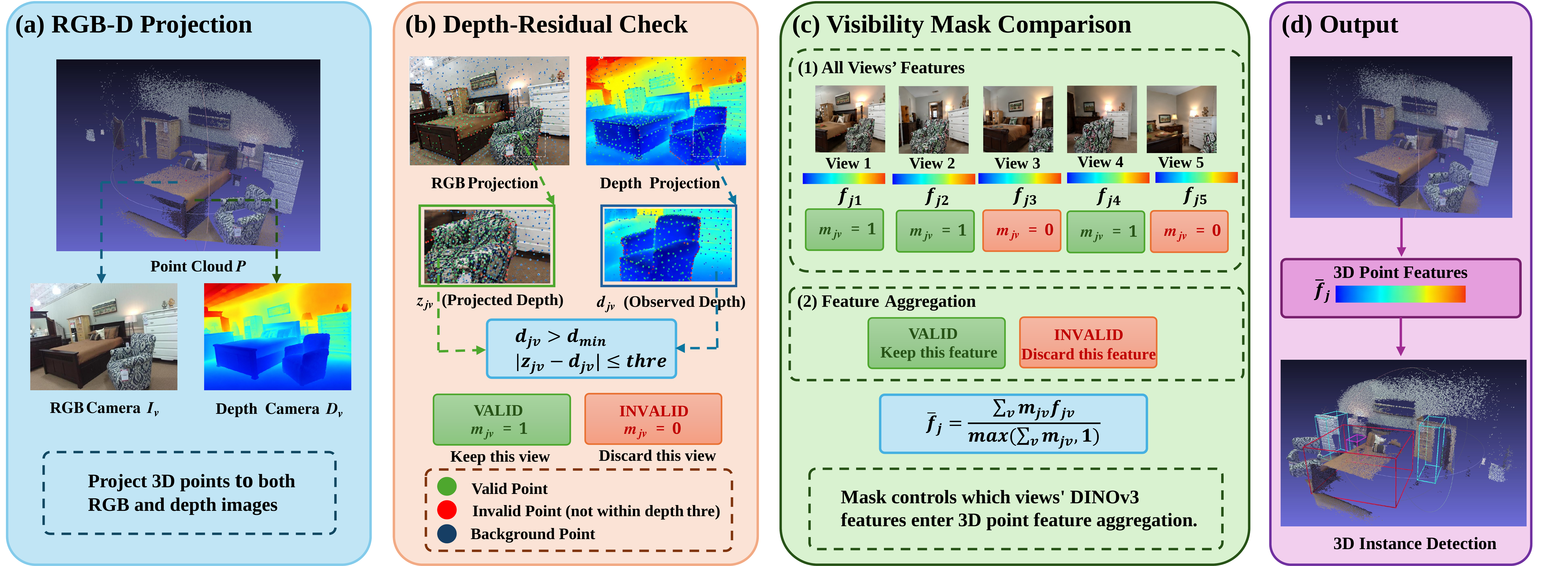}}
\newcommand{\realsetupimage}{\includegraphics[width=0.98\columnwidth]{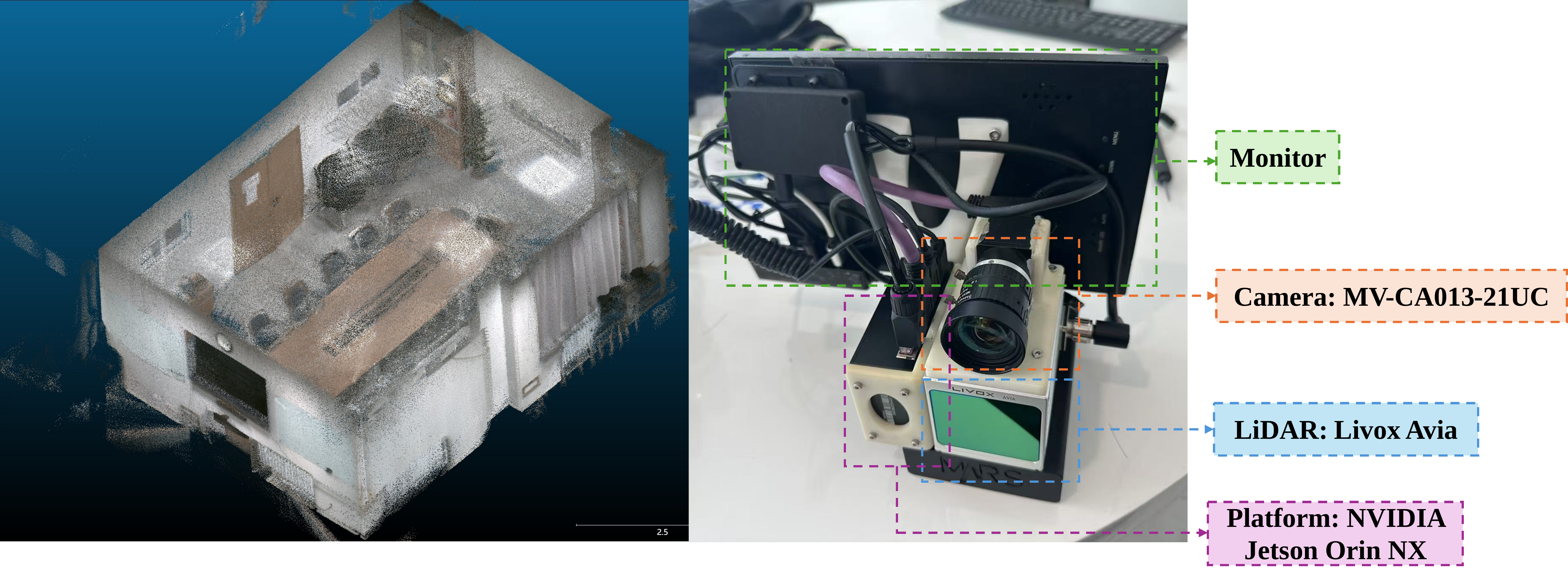}}
\newcommand{\realcomparisonimage}{\includegraphics[width=0.98\columnwidth]{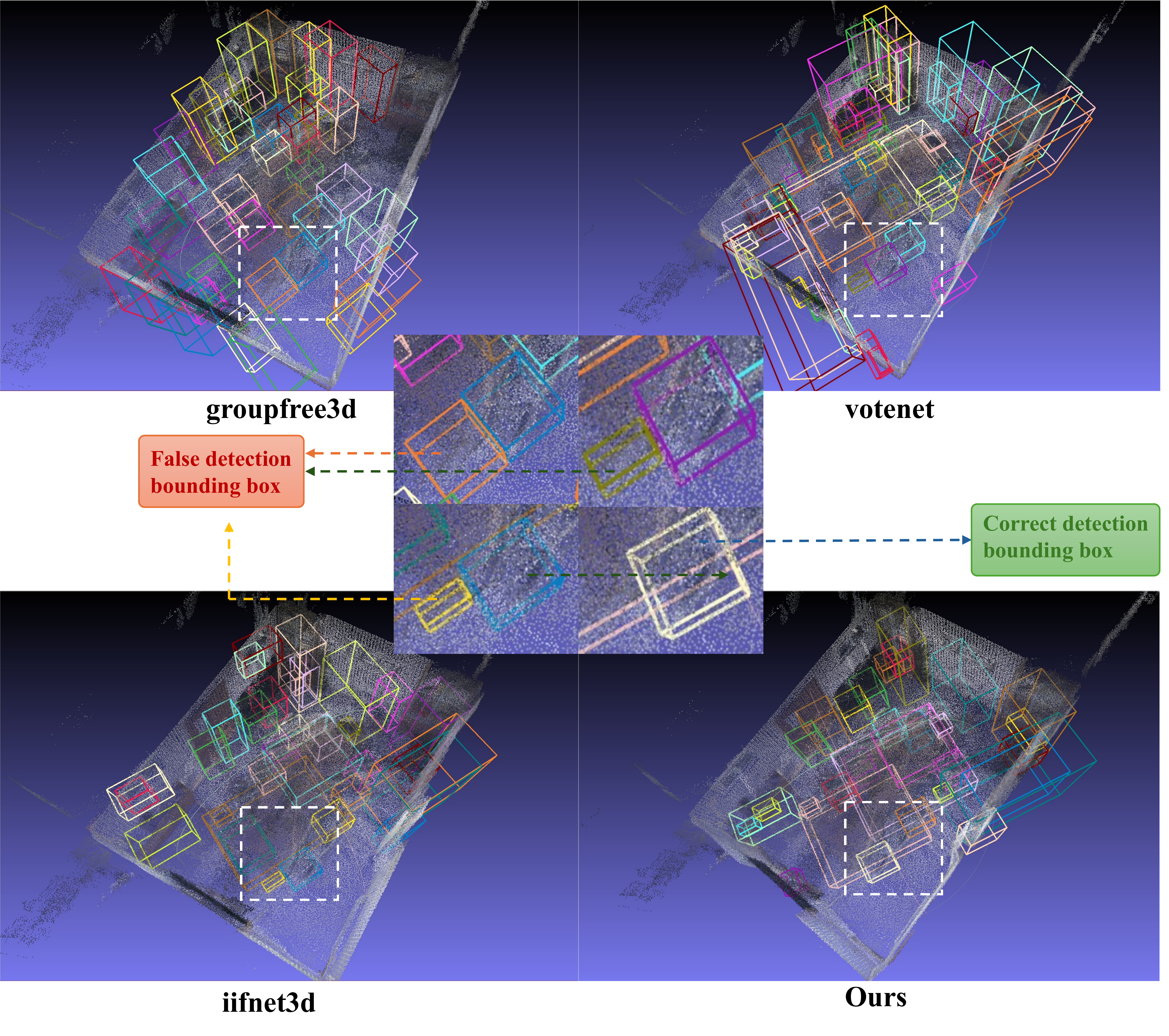}}

\newcounter{algorithm}
\newenvironment{tablehere}{%
  \par\addvspace{\intextsep}\noindent
  \begin{minipage}{\columnwidth}
  \def\@captype{table}\centering
}{%
  \end{minipage}\par\addvspace{\intextsep}
}
\newenvironment{figurehere}{%
  \par\addvspace{\intextsep}\noindent
  \begin{minipage}{\columnwidth}
  \def\@captype{figure}\centering
}{%
  \end{minipage}\par\addvspace{\intextsep}
}

\newcommand{\widecaptiontable}[1]{\def\@captype{table}\caption{#1}}
\newcommand{\widecaptionfigure}[1]{\def\@captype{figure}\caption{#1}}
\makeatother
\title{\LARGE \bf DIFTA-3D: Depth-Consistent Instance-Level Feature Transfer and Adaptation of DINOv3 for 3D Detection}

\author{Linman Wang$^{1}$, ZiFei Zhang$^{1}$, Chunran Zheng$^{2}$, Xiwang Dong$^{1}$, Jiarong Lin$^{1,*}$%
\thanks{$^{1}$Beihang University.}%
\thanks{$^{2}$The University of Hong Kong.}%
\thanks{$^{*}$Corresponding author: Jiarong Lin, {\tt zivlin@connect.hku.hk}.}%
}

\begin{document}
\maketitle
\thispagestyle{empty}
\pagestyle{empty}

\begin{abstract}
RGB-D 3D instance detectors benefit from visual semantics, but the task-specific Faster R-CNN/ResNet branch used by IIFNet3D couples feature extraction to a separately trained 2D detector and its image-domain labels.
{Replacing that branch with a frozen vision foundation model removes this task-specific dependency, but may introduce occlusion noise and a mismatch between patch features and geometry-aware detection features.}
In this work, we {investigate this replacement through an adaptation of DINOv3 to the instance-level fusion pipeline of IIFNet3D}.
{At the core of our approach is a depth-consistent feature pipeline that} {projects scene points into calibrated RGB-D frames, applies a metric depth-residual check, averages the accepted DINOv3 features into an offline point cache, and aggregates the cached features inside proposal-aligned RoI grids}. 
The geometric and bidirectional instance-fusion paths are {preserved}, while Conservative VAID is {evaluated as a low-strength, support-weighted semantic distillation recipe applied only to positive RoIs}.
{We conduct extensive evaluations on ScanNetV2 to assess the proposed transfer recipes.}
{On ScanNetV2, our DINOv3 control achieves mAP scores of $76.15$ and $60.93$ at IoU thresholds of $0.25$ and $0.50$, respectively.}
{The Conservative VAID setting {achieves mAP scores of} $76.59$ and $62.16$, {corresponding to numerical gains of} $0.44$ and $1.23$ points over the control, respectively, in this checkpoint-level recipe comparison.}
The reported IIFNet3D result of $75.7/63.8$ is used only as an external reference because the visual branch and processing protocol differ.
{{Accordingly, we interpret these results as evidence for a controlled transfer recipe rather than as a causal estimate of the individual contributions of VAID or depth filtering.}}
\end{abstract}

\section{INTRODUCTION}
Three-dimensional instance detection from RGB-D observations is central to indoor scene understanding and embodied robotics \cite{scannet,sunrgbd}.
A reliable detector must combine geometric evidence from point clouds with visual semantics that distinguish objects with similar shapes or limited geometric support.
Vision foundation models offer a promising source of transferable semantics because their frozen representations can be reused across image domains without training a task-specific image detector \cite{dinov3}.
Yet a general-purpose patch representation is not inherently aligned with a 3D proposal, an occluded image measurement, or a geometry-sensitive detection head.
The key challenge is therefore to adapt foundation-model features for instance-level 3D reasoning while preserving the geometric representation that supports detection.

Existing RGB-D 3D detectors exploit point-cloud geometry through voting, grouping, or point-to-image fusion; some also use image features from a task-specific 2D detector \cite{votenet,pointfusion,imvotenet,iifnet3d}. Instance-level fusion is attractive because each proposal can aggregate visual evidence before the 3D detection head. The original IIFNet3D branch has a useful detection prior, but it requires a dedicated 2D detector and does not provide the broad, frozen patch representation provided by DINOv3.
Introducing DINOv3 raises three unresolved issues. First, projecting a 3D sample into an RGB frame may retrieve a background patch when the sample is occluded. Second, a fixed ROI grid can provide unstable coverage for thin, boundary, or sparsely observed objects. Third, the semantic distribution of a frozen foundation model differs from that of the detection-specific visual features used to design the fusion module. Our controlled experiments show that direct cross-attention, proposal/ROI gating, residual fusion, and strong visual distillation can all reduce performance. These observations motivate explicit visibility control and task-constrained transfer rather than direct feature substitution.

We address these issues with DIFTA-3D, a depth-consistent instance-level transfer of DINOv3 within IIFNet3D.
The method replaces the task-specific visual encoder while retaining the geometric branch, the proposal ROI grid with RoI-Conv pooling, and the bidirectional GGF/SGF fusion path.
{For each scene point projected into a calibrated frame, we compare the camera-space depth with the depth image and retain the visual sample only when {the discrepancy between them} falls below a predefined tolerance.}
Accepted samples are averaged into a 384-dimensional point cache, and unobserved points are set to zero before proposal pooling.
On top of this construction, Conservative VAID uses detached DINOv3 features as teachers and applies low-strength cosine distillation only to positive RoIs, with a feature-energy support weight.
The central insight is to expose DINOv3's broad semantics through depth-consistent instance evidence while constraining the transfer to protect the detector's geometry, rather than treating a foundation-model feature map as a drop-in replacement.

Our ScanNetV2 checkpoint comparison illustrates this positioning. The DINOv3 control obtains $76.15$ mAP@0.25 and $60.93$ mAP@0.50, whereas Conservative VAID obtains $76.59$ and $62.16$, respectively. Thus, the observed recipe-level difference is $0.44$ points at mAP@0.25 and $1.23$ points at mAP@0.50. These differences combine continuation optimization with the auxiliary recipe and are not an isolated estimate of VAID. The reported IIFNet3D result of $75.7/63.8$ is included only as an external reference because its visual branch and processing protocol differ from those used in our experiments.

Our contributions are summarized as follows:
\vspace{-0.15cm}
\textbf{\begin{itemize}
    \item {We identify the {feature-distribution and visibility mismatches} that arise when a task-specific 2D visual branch is replaced by a frozen DINOv3 encoder in an instance-level RGB-D detector.}
    \item {We construct depth-consistent DINOv3 instance features {using calibrated depth-residual filtering and proposal-aligned RoI aggregation}, and evaluate Conservative VAID as a low-strength, support-weighted semantic regularizer for positive RoIs.}
    \item {We provide a controlled ScanNetV2 evaluation and a {cross-dataset protocol analysis}, reporting category-level behavior, high-IoU performance, and negative results for simpler transfer alternatives while separating observed recipe effects from {unverified component-level causal effects}.}
\end{itemize}}

\section{RELATED WORK}

{Point-based detectors extract local geometric evidence directly from irregular point sets, whereas voting-based methods generate object-center hypotheses from surface points \cite{votenet,h3dnet}. Transformer-based and sparse-voxel detectors provide complementary mechanisms for modeling long-range context and operating on sparse scenes \cite{groupfree3d,threedetr,gsdn}. These methods establish strong geometric foundations for indoor detection, but thin or weakly sampled surfaces can remain ambiguous when appearance is needed to distinguish instances.}

{Multimodal detectors combine point-cloud geometry with RGB semantics through point-to-pixel projection, voxel lifting, or cross-modal attention \cite{pointfusion,imvotenet,tr3d}. IIFNet3D performs proposal-level instance-to-instance fusion with geometry-guided and semantics-guided attention \cite{iifnet3d}. We follow this instance-level design while studying a distinct question: how to replace its task-specific visual branch with DINOv3. This replacement changes both the feature distribution and the reliability of projected evidence. We therefore retain the original fusion path and redesign the visual feature construction and adaptation protocol around depth consistency.}

{Self-supervised vision encoders such as DINO\cite{dino}, DINOv2\cite{dinov2}, and DINOv3\cite{dinov3} provide patch-level representations that transfer across image domains. Their semantic breadth is useful when a task-specific 2D detector is unavailable. However, patch features are not inherently aware of 3D visibility, proposal boundaries, or the optimization behavior of a geometry-trained detector. In our setting, DINOv3 is consequently used as a frozen teacher, and its evidence is adapted at the instance level rather than used to replace the geometric representation.}

{Depth consistency and occlusion reasoning have been used to constrain image-to-3D feature lifting \cite{tr3d,transfusion}. Our setting differs in that visibility is evaluated for proposal ROI samples across calibrated views and then used to construct an instance feature. This formulation supports explicit invalid-view handling and provides a basis for analyzing hard rejection and effective coverage, including failure modes involving thin and boundary objects. Together, these distinctions motivate our depth-consistent adaptation of DINOv3 within an instance-level 3D detector.}

{\section{METHODOLOGY}}

\subsection{Overview}
\begin{figure*}[!t]
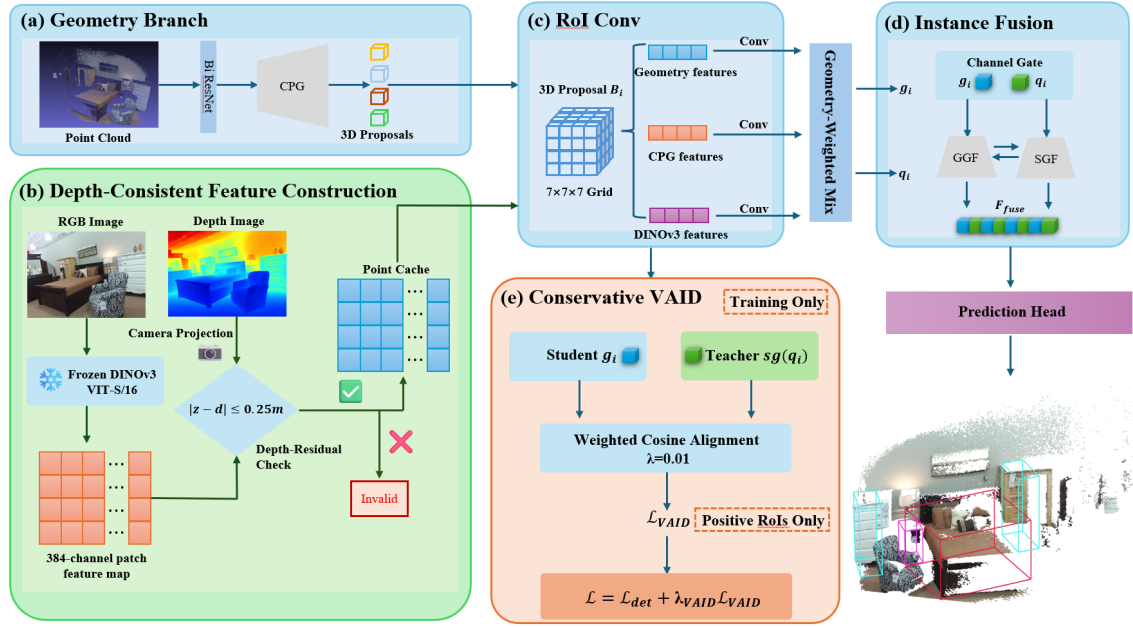

\centering
\noindent\makebox[\linewidth][c]{\pipelineimagefull}\par
\vspace{-0.5cm}
\caption{\textbf{Overview of DIFTA-3D.} The geometric path generates 3D proposals, while projected DINOv3 features are filtered by a depth-residual check before RoI aggregation and bidirectional instance fusion. Conservative VAID uses detached teacher features, visibility weighting, and positive-RoI supervision.}
\vspace{-0.2cm}
\label{fig:pipeline}
\end{figure*}

{DIFTA-3D replaces the task-specific visual branch of IIFNet3D with depth-consistent DINOv3 evidence while preserving the geometric proposal path and the instance-fusion blocks. The framework {comprises} three stages: depth-consistent feature construction, proposal-aligned visual aggregation, and conservative semantic adaptation, as illustrated in Fig.~\ref{fig:pipeline}.}

{Let $P$ denote the input point cloud and let $\{(I_v,D_v,\Pi_v)\}_{v=1}^{V}$ denote calibrated RGB images, depth images, and camera projection matrices. A geometric encoder extracts point features, and the coarse proposal generator (CPG) produces a set of 3D proposals $\{B_i\}_{i=1}^{N}$. For each proposal, the visual branch constructs an instance descriptor from DINOv3, the bidirectional instance modules fuse it with proposal geometry, and the detection head performs classification and box regression. The following subsections define these stages in detail.}

\subsection{Depth-Consistent DINOv3 Features}
{The principal ScanNet experiments use an offline point-feature cache.
The frozen DINOv3 ViT-S/16 model produces a 384-channel patch map $F_v=\Phi(I_v)$.
For an aligned scene point $p_j$, let $\hat p_j=[p_j^\top,1]^\top$, and let $M_v^c,M_v^d$ denote the calibrated color and depth projection matrices, including the inverse scene alignment.
For sensor $s\in\{c,d\}$, the projection and feature sampling operations are defined as follows.}

\begin{equation}
\begin{IEEEeqnarraybox}[][c]{rCl}
h^s_{j,v}&=&M_v^s\hat p_j,\quad z^s_{j,v}=h^s_{j,v,3},\\
u^s_{j,v}&=&h^s_{j,v,1:2}/z^s_{j,v},\\
f_{j,v}&=&\mathrm{Bilinear}\bigl(F_v,\mathrm{round}(u^c_{j,v})/16\bigr),
\end{IEEEeqnarraybox}
\label{eq:projection}
\end{equation}
{where} {$u^s_{j,v}$ is a two-dimensional pixel coordinate, and $z^s_{j,v}$ is camera-space depth. Color coordinates are rounded before sampling, as in the cache builder. Let $a_{j,v}$ require a positive color depth and an in-bounds rounded color coordinate. With $z_{j,v}=z^d_{j,v}$ and depth observation $d_{j,v}$, the mask is}
\begin{equation}
 m_{j,v}=a_{j,v}\,\mathbf{1}[d_{j,v}>0.05\,\mathrm{m}]
 \,\mathbf{1}[|z_{j,v}-d_{j,v}|\leq \tau_d],
\label{eq:visibility}
\end{equation}
{where} {$\tau_d$ denotes the depth-residual threshold, {$\mathbf{1}[\cdot]$ denotes the indicator function, which equals $1$ when its condition is satisfied and $0$ otherwise}. Raw depth images are converted to metric units, and projected depth coordinates are rounded and clamped to the valid image range.
If a depth image is unavailable, the cache builder falls back to the front-facing and color-in-bounds test; strict occlusion filtering therefore requires a valid depth input.
With $n_j=\sum_v m_{j,v}$, the cached feature is:}
\begin{equation}
\bar f_j=\frac{\sum_v m_{j,v}f_{j,v}}{\max(n_j,1)}.
\label{eq:cache}
\end{equation}

\begin{figure*}[!t]
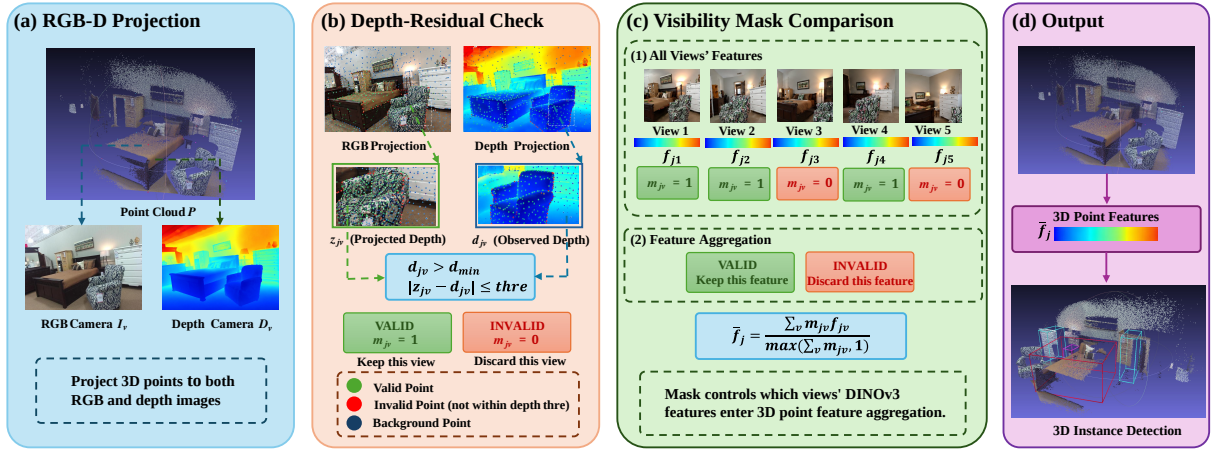

\centering
\noindent\makebox[\linewidth][c]{\depthimagefull}\par
\caption{\textbf{Depth-consistent point-feature construction.} Each 3D point is projected independently into calibrated RGB and depth cameras.
Frozen DINOv3 patches provide the color feature, while the metric depth sample checks $d>0.05\,\mathrm{m}$ and the residual $|z-d|\leq0.25\,\mathrm{m}$; foreground observations are retained, and occluded background observations are rejected.
The comparison shows how depth filtering prevents contaminated features from being aggregated.
Accepted observations are averaged into a 384-dimensional offline point cache, with an explicit zero feature assigned when no view is valid.
A proposal-aligned $7\times7\times7$ grid queries this cache, and RoI-Conv pooling produces the instance descriptor $q_i$.}
\vspace{-0.5cm}
\label{fig:depth}
\end{figure*}

{Points with no accepted observation retain a zero feature.
This is a point-level cache rule; subsequent learned pooling can still transform the resulting descriptor.
During training and evaluation, the cache is loaded together with the point cloud and queried within each proposal.
The depth projection and invalid-observation logic are summarized in Fig.~\ref{fig:depth}.}

\subsection{Instance-Level Fusion and Conservative VAID}
\subsubsection{Proposal Generation and RoI-Conv Aggregation.}
The inherited geometric stream voxelizes XYZRGB observations and extracts 64-channel BiResNet features.
The CPG encoder combines local and superpoint context, and its prediction head produces class scores, centerness, and box parameters for axis-aligned ScanNet bounding boxes.
The refinement stage samples a fixed set of proposals per scene.
These components are kept unchanged in the DINOv3 experiments.

For an axis-aligned ScanNet proposal with center $b_i$ and dimensions $\ell_i$, we construct a proposal-aligned grid of cell-center queries.
With $k\in\{0,\ldots,6\}^3$, the grid and visual pooling operations are defined as follows:
\begin{equation}
\begin{IEEEeqnarraybox}[][c]{rCl}
x_{i,k}&=&b_i+\ell_i\odot\bigl((k+0.51)/7-0.51\bigr),\\
q_i&=&\mathcal{R}_7\bigl(\mathcal{R}_5(\mathcal{S},\mathcal{G}_i)\bigr),
\end{IEEEeqnarraybox}
\label{eq:roi}
\end{equation}
{where $\mathcal{G}_i=\{x_{i,k}\}$ and $\mathcal{S}$ is the sparse tensor obtained by voxelizing $\{(p_j,\bar f_j)\}$.
The operator $\mathcal{R}_5$ queries voxelized grid locations with a kernel-size-five sparse convolution, batch normalization, and ELU.
The operator $\mathcal{R}_7$ reorganizes these features on the local $7^3$ grid and applies a kernel-size-seven convolution and batch normalization at its center.}
This {produces} a 128-dimensional descriptor.
The same pooling operation is applied to the 64-channel geometric decoder and 390-channel CPG streams; their descriptors are combined using a learned scalar sigmoid weight to obtain $g_i$.

The cache construction and proposal pooling follow the sequence specified by Eqs.~(\ref{eq:projection})--(\ref{eq:roi}); the resulting descriptors are then passed to instance fusion.

\subsubsection{Bidirectional instance fusion.}
The pooled descriptors interact through the geometry-guided fusion (GGF) and semantics-guided fusion (SGF) blocks.
The retained channel-wise gate is applied before cross-attention:
\begin{equation}
\tilde g_i=\gamma_i\odot g_i+(1-\gamma_i)\odot q_i,~~
\gamma_i=\sigma\!\left(\mathrm{MLP}([g_i;q_i])\right)
\label{eq:gate}
\end{equation}
This is the pre-attention gate in the current DINOv3 configuration; the proposal/ROI gate evaluated as P0 is a separate ablation.
Let $E_q$ and $E_k$ be learned MLPs that encode the sine-transformed box center, dimensions, and volume, and let $E_c$ encode the proposal center.
{{Here, $\mathcal{A}(X,Y)\equiv\mathrm{MHA}(X,Y,Y)$ denotes single-head scaled dot-product attention, $\mathrm{LN}(\cdot)$ denotes Layer Normalization, and the second argument supplies both the key and value; the implemented two directions can be written as}}
\begin{equation}
\scriptstyle
\begin{array}{c}
H_g=\mathrm{LN}\bigl(\mathcal{A}(\tilde G+E_q,Q+E_k)+\tilde G+E_q\bigr),\\\\
H_q=\mathrm{LN}\bigl(\mathcal{A}(Q+E_c,\tilde G+E_c)+Q+E_c\bigr),\\\\[-1pt]
H=H_g+H_q.
\end{array}
\label{eq:fusion}
\end{equation}
where $\tilde G$ and $Q$ stack the proposal descriptors within each scene.
Each attention block uses 128 channels, one head, a dropout rate of 0.1, and a residual LayerNorm.
The output is additive, consistent with the reference implementation; it is not formed by concatenation.

\subsubsection{Conservative VAID.}
Conservative VAID treats the pooled geometric descriptor $s_i=g_i$ as the student and the detached visual descriptor $t_i=\mathrm{sg}(q_i)$ as the teacher.
The offline cache does not store a calibrated proposal visibility count.
We therefore use a detached feature-energy proxy.
For $e_i=\|t_i\|_2$ and local mean $\bar e$, the implementation computes
\begin{equation}
r_i=\mathrm{clip}\!\left(\frac{e_i}{2\max(\bar e,10^{-6})},0,1\right),~ c_i=0.25+0.75r_i.
\label{eq:support}
\end{equation}

Thus, $c_i$ is a support proxy rather than a measured visibility fraction.
Here, $\bar e$ is the mean of $e_i$ over all RoIs in the current batch.
For the positive RoIs $\mathcal{P}$ selected by the detector's regression-valid mask, we define the auxiliary objective as
\begin{equation}
\mathcal{L}_{\mathrm{vaid}}=\frac{\sum_{i\in\mathcal{P}}c_i\left(1-\cos(s_i,\mathrm{sg}(t_i))\right)}{\max\left(\sum_{i\in\mathcal{P}}c_i,10^{-6}\right)},
\label{eq:vaid}
\end{equation}
where cosine similarity is computed between normalized features, and $\mathrm{sg}$ denotes stop-gradient.
If no positive RoI is present, the auxiliary loss is set to zero.
DINOv3 remains frozen, and the auxiliary term introduces neither a second encoder pass nor additional adapter parameters.
The total objective is
\begin{equation}
\mathcal{L}=\mathcal{L}_{\mathrm{det}}+\lambda_{\mathrm{vaid}}\mathcal{L}_{\mathrm{vaid}}.
\label{eq:objective}
\end{equation}

The reported conservative setting uses an adaptation learning rate of $1\times10^{-5}$ and $\lambda_{\mathrm{vaid}}=0.01$.
The low loss weight and positive-RoI restriction make the auxiliary term serve as a semantic-alignment regularizer rather than a replacement for the detection objective.

\subsubsection{Detection Supervision and Inference}
The detector retains supervision at both stages.
The CPG objective combines focal classification, centerness, distance-IoU box regression, and Smooth-L1 voting losses.
The refinement objective combines classification, Smooth-L1 residual regression, and distance-IoU losses with weights $1$, $0.5$, and $1$, respectively; the refinement and CPG losses are summed to define $\mathcal{L}_{\mathrm{det}}$.
The refinement target uses the inherited regression-valid mask, and a previously negative proposal becomes positive when its maximum ground-truth IoU is at least $0.3$.
At inference, VAID supervision is omitted; the detector uses standard refinement decoding and NMS, with a pre-NMS limit of $1,000$ candidates and an IoU threshold of $0.5$.

\section{EXPERIMENTS}
\begin{table*}[!t]
\caption{\textbf{ScanNetV2 category-wise detection results (mAP@0.25).} Original-paper rows are external references; DINOv3 rows use the corrected in-house evaluator. Bold entries mark the larger of the two reported DINOv3 values for each category.}
\label{tab:main}
\centering\tablebodyfont
\setlength{\tabcolsep}{2.00pt}
\renewcommand{\arraystretch}{1.02}
\noindent\makebox[\textwidth][c]{%
\begin{tabular}{@{}lrrrrrrrrrrrrrrrrrrr@{}}
\hline
Method & cab & bed & chr & sofa & tbl & door & wnd & bks & pic & ctr & desk & crt & frg & shw & tol & snk & bth & ofn & mAP \\
\hline
\multicolumn{20}{@{}l}{\itshape Point Cloud-Driven} \\
\hline
GSDN~\cite{gsdn} & 41.6 & 82.5 & 92.1 & 87.0 & 61.1 & 42.4 & 40.7 & 51.1 & 10.2 & 64.2 & 71.1 & 54.9 & 40.0 & 70.5 & 99.9 & 75.5 & \textbf{93.2} & 53.1 & 62.8 \\
VoteNet~\cite{votenet} & 47.7 & 88.7 & 89.5 & 89.3 & 62.1 & 54.1 & 40.8 & 54.3 & 12.0 & 63.9 & 69.4 & 52.0 & 52.5 & 73.3 & 95.9 & 52.0 & 92.5 & 41.4 & 62.9 \\
Pointformer~\cite{pointformer} & 46.7 & 88.4 & 90.5 & 88.7 & 65.7 & 55.0 & 47.7 & 55.8 & 18.0 & 63.8 & 69.1 & 55.4 & 48.5 & 66.2 & 98.9 & 61.5 & 86.7 & 47.4 & 64.1 \\
MLCNet~\cite{mlcnet} & 42.5 & 88.5 & 90.0 & 87.4 & 63.5 & 56.9 & 47.0 & 56.9 & 11.9 & 63.9 & 76.1 & 56.7 & 60.9 & 65.9 & 98.3 & 59.2 & 87.2 & 47.9 & 64.5 \\
BRNet~\cite{brnet} & 49.9 & 88.3 & 91.9 & 86.9 & 69.3 & 59.2 & 45.9 & 52.1 & 15.3 & 72.0 & 76.8 & 57.1 & 60.4 & 73.6 & 93.8 & 58.8 & 92.2 & 47.1 & 66.1 \\
H3DNet~\cite{h3dnet} & 49.4 & 88.6 & 91.8 & 90.2 & 64.9 & 61.0 & 51.9 & 54.9 & 18.6 & 62.0 & 75.9 & 57.3 & 57.2 & 75.3 & 97.9 & 67.4 & 92.5 & 53.6 & 67.2 \\
GroupFree3D~\cite{groupfree3d} & 52.1 & 91.9 & 93.6 & 88.0 & 70.7 & 60.7 & 53.7 & 62.4 & 16.1 & 58.5 & 80.9 & 67.9 & 47.0 & 76.3 & 99.6 & 72.0 & 95.3 & 56.4 & 69.1 \\
SCGNet~\cite{scgnet} & -- & -- & -- & -- & -- & -- & -- & -- & -- & -- & -- & -- & -- & -- & -- & -- & -- & -- & 69.1 \\
Objformer~\cite{objformer} & 55.4 & 88.7 & 93.4 & 87.2 & 74.1 & 61.3 & 57.3 & 55.5 & 17.9 & 67.4 & \textbf{85.1} & \textbf{74.4} & 52.0 & 79.8 & 97.5 & 71.8 & 88.2 & 57.5 & 70.3 \\
FCAF3D~\cite{fcaf3d} & 57.2 & 87.0 & 95.0 & 92.3 & 70.3 & 61.1 & 60.2 & 64.5 & 29.9 & 64.3 & 71.5 & 60.1 & 52.4 & \textbf{83.9} & 99.9 & 84.7 & 86.6 & 65.4 & 71.5 \\
TR3D~\cite{tr3d} & -- & -- & -- & -- & -- & -- & -- & -- & -- & -- & -- & -- & -- & -- & -- & -- & -- & -- & 72.9 \\
DLLA~\cite{dlla} & 56.0 & 86.8 & \textbf{96.3} & 91.5 & 74.8 & 63.2 & 57.2 & 65.0 & 32.7 & 75.8 & 82.5 & 57.9 & 60.7 & 83.7 & 99.8 & 80.2 & 90.2 & 64.8 & 73.8 \\
SPGroup3D~\cite{spgroup3d} & 58.0 & 88.2 & 94.2 & \textbf{93.0} & 73.4 & 68.4 & 65.9 & 66.9 & 39.3 & 72.5 & 79.6 & 64.2 & 64.0 & 79.6 & 99.8 & 77.3 & 90.2 & 62.2 & 74.3 \\
CAGroup3D~\cite{cagroup3d} & 60.4 & \textbf{93.0} & 95.3 & 92.3 & 70.0 & 68.0 & 63.6 & 67.3 & 40.7 & 77.0 & 83.9 & 69.4 & 65.7 & 73.0 & 100.0 & 79.7 & 87.0 & 66.1 & 75.1 \\
\hline
\multicolumn{20}{@{}l}{\itshape Multi-modal} \\
\hline
MFFVoteNet~\cite{mffvotenet} & 40.5 & 89.0 & 89.1 & 85.5 & 64.4 & 57.6 & 49.8 & 58.9 & 14.4 & 63.4 & 69.8 & 51.6 & 51.6 & 71.2 & 97.3 & 59.5 & 91.4 & 45.5 & 63.9 \\
PiMAE~\cite{pimae} & -- & -- & -- & -- & -- & -- & -- & -- & -- & -- & -- & -- & -- & -- & -- & -- & -- & -- & 67.6 \\
TokenFusion~\cite{tokenfusion} & -- & -- & -- & -- & -- & -- & -- & -- & -- & -- & -- & -- & -- & -- & -- & -- & -- & -- & 69.8 \\
SPGroup3D+FF$^{\dagger}$ & \textbf{62.4} & 89.9 & 94.3 & 92.0 & 73.3 & 69.7 & 68.3 & \textbf{73.6} & 44.1 & 65.1 & 80.2 & 62.9 & 64.9 & 70.6 & 100.0 & 78.2 & 91.9 & 63.8 & 74.7 \\
IIFNet3D (ext.) & 62.1 & 90.2 & 94.9 & 92.7 & \textbf{77.0} & \textbf{70.5} & 68.7 & 68.1 & \textbf{49.5} & 63.0 & 82.5 & 65.4 & 64.9 & 79.7 & 100.0 & 77.5 & 91.4 & 65.3 & 75.7 \\
\hline
\multicolumn{20}{@{}l}{\itshape DINOv3 adaptation (current evaluator)} \\
\hline
DINOv3 control & 62.30 & 83.25 & 89.93 & 87.09 & 70.59 & 64.99 & 71.21 & 70.91 & 47.37 & 80.71 & 81.57 & 72.43 & 73.64 & 71.78 & 100.00 & 85.31 & 89.35 & 68.22 & 76.15 \\
Conservative VAID & 61.54 & 83.19 & 90.01 & 86.01 & 70.51 & 65.11 & \textbf{72.74} & 72.35 & 47.38 & \textbf{81.32} & 81.44 & 73.63 & \textbf{74.28} & 72.82 & \textbf{100.00} & \textbf{86.32} & 90.03 & \textbf{69.98} & \textbf{76.59} \\
\hline
\end{tabular}
}%
\par\vspace{1mm}
\noindent\parbox[t]{\textwidth}{\footnotesize $^{\dagger}$ Early-stage fusion in the original paper. Class abbreviations follow the ScanNetV2 labels: cab (cabinet), chr (chair), tbl (table), wnd (window), bks (bookshelf), pic (picture), ctr (counter), crt (curtain), frg (refrigerator), shw (shower), tol (toilet), snk (sink), bth (bathtub), and ofn (other furniture). External and DINOv3 rows use different visual branches and processing protocols. Bold marks the larger of the two reported DINOv3 values in each column; external rows are not used for bolding.}
\vspace{-0.5cm}
\end{table*}
{\subsection{Experimental Setup}}
We evaluate ScanNetV2 {across} its 18 classes at IoU thresholds of $0.25$ and $0.50$.
The principal control uses the depth-consistent offline DINOv3 cache and the same detector configuration as Conservative VAID, but without the auxiliary loss.
The name ``times=8'' refers to the RepeatDataset training multiplier, not to eight RGB frames; each cached point feature may aggregate features from all frames available for its scene.
Training uses four-GPU distributed data parallelism without validation during training, and evaluation follows the corrected four-GPU protocol.
The reported IIFNet3D values serve as external references rather than same-protocol baselines because the visual branch and processing details differ.

{All comparisons in this paper use the protocols and evaluators described above.
Quantities that are unavailable in an external source are omitted rather than used to support a numerical claim.
For provenance, the in-house runs use seed $0$, deterministic data-loader settings, and a fixed checkpoint rule with no {best-checkpoint selection}: each VAID recipe is evaluated after a one-epoch continuation from the archived control checkpoint.
The evaluator records the ordered AP vector, IoU threshold, checkpoint, and seed before computing mAP from unrounded values; the reproducibility artifact should expose these records, the evaluator version, configuration, and exact commands.}

\subsection{Datasets and Metrics}
\subsubsection{ScanNetV2}
ScanNetV2 contains 1,513 reconstructed indoor scenes, with 1,201 scenes for training and 312 scenes for validation.
We follow the 18-category protocol used by the IIFNet3D implementation and report mean average precision at 3D IoU thresholds of $0.25$ and $0.50$.
The corrected in-house evaluator uses axis-aligned corner IoU for the ScanNet boxes and the 11-point AP calculation implemented in the archived evaluation script.
The evaluator computes each mAP from the unrounded 18-class AP vector in the same log and rounds the final scalar to two decimals; the class entries printed in Table~I are independently rounded. The Table~I rows were regenerated from the archived epoch-17 control and epoch-1 Conservative VAID logs.

\subsubsection{SUN-RGBD}
SUN-RGBD contains $10,335$ indoor RGB-D images, with $5,285$ training images and $5,050$ validation images.
Following the standard $10$-category protocol, we report mAP at 3D IoU thresholds of $0.25$ and $0.50$.
The corrected evaluator recomputes rotated 3D IoU from box corners and integrates the precision--recall curve over all recall changes.
Thus, the AP implementation is documented separately for the two datasets rather than being assumed to be identical.
The original IIFNet3D paper reports only mAP at a 3D IoU threshold of $0.25$ for its SUN-RGBD reference row.

\vspace{-0.5cm}
\begin{tablehere}
\caption{\textbf{VAID training-recipe comparison on ScanNetV2.}}
\label{tab:scan_summary}
\centering\tablebodyfont
\renewcommand{\arraystretch}{1.0}
\setlength{\tabcolsep}{2.2pt}
\begin{tabular}{@{}l@{\hspace{0.35em}}cc@{\hspace{0.35em}}lcc@{}}
\hline
Configuration & LR & $\lambda$ & Support & AP25 & AP50 \\
\hline
\multicolumn{6}{l}{\itshape External reference} \\
\hline
IIFNet3D & -- & -- & -- & 75.70 & \textbf{63.80} \\
\hline
\multicolumn{6}{l}{\itshape DINOv3 control and training recipes} \\
\hline
DINOv3 control & -- & 0 & None & 76.15 & 60.93 \\
VAID, uniform & $10^{-3}$ & 0.05 & Uniform & 70.99 & 57.80 \\
VAID, energy-weighted & $10^{-5}$ & 0.01 & Energy & \textbf{76.59} & 62.16 \\
\hline
\end{tabular}
\par\vspace{0.5mm}
\noindent\parbox{\linewidth}{\footnotesize LR and $\lambda$ refer to VAID fine-tuning.
{``Uniform'' and ``Energy''} denote uniform and detached feature-energy support weighting on positive RoIs.
AP25/AP50 denote mAP at IoU thresholds of {$0.25$ and $0.50$}, respectively. The two VAID rows jointly change {the continuation learning rate}, loss weight, and support rule; they are not a one-factor ablation.
Bold marks the best reported DINOv3 result.}
\end{tablehere}

\subsection{Implementation Details}
The detector is implemented with the MMDetection3D framework.
The geometric stream uses the BiResNet-based sparse 3D backbone with a voxel size of $0.02\,\mathrm{m}$.
The principal ScanNetV2 control uses AdamW with an initial learning rate of $1\times10^{-3}$, weight decay of $1\times10^{-4}$, and learning-rate decays at epochs 9, 12, and 15 for 20 epochs.
The distributed batch uses four samples per GPU.
DINOv3 is a frozen ViT-S/16 encoder with 384-channel cached patch features; the RoI-Conv path maps each stream to 128 channels.
The principal cache uses the depth-consistency threshold specified above, and the training configuration uses a RepeatDataset multiplier of eight.
The RoI head samples a $7\times7\times7$ grid, retains 128 proposals per scene, and uses kernel sizes of five and seven in its two sparse pooling stages.
Conservative VAID is initialized from the archived control checkpoint and fine-tuned for the reported screening run with a learning rate of $1\times10^{-5}$ and a loss weight of $0.01$.
The packed train and validation caches occupy $171.9480$ decimal GB, and the checkpoint used for the reported ScanNetV2 rows contains $47.5587$ million parameters. The logged peak memory is $57{,}570$ MB for the control and $68{,}876$ MB for Conservative VAID; an inference-throughput benchmark was not archived and is therefore not reported.

The original IIFNet3D values in Table~\ref{tab:main} are external references: its task-specific Faster R-CNN branch and processing protocol were not rerun here. The DINOv3 rows are produced by the current evaluator, and unavailable external quantities are omitted.

\subsection{Evaluation Results}
Table~\ref{tab:scan_summary} compares the reference detector, the DINOv3 control, and two VAID training recipes.
Both VAID variants start from the same control checkpoint and are evaluated after one additional training epoch, using the same depth-checked cache, proposal pooling, and corrected evaluator.
Because the recipes jointly change the continuation learning rate, auxiliary-loss weight, and support weighting, this table is a recipe comparison rather than a causal isolation of VAID.
The tables use the archived control checkpoint and the first Conservative VAID continuation checkpoint; no multi-seed or best-of-epoch selection is claimed.
Accordingly, the rows should be read as checkpoint-level observations under a fixed protocol, not as estimates of a population mean or a statistically stable method effect.

\begin{figure*}[!t]
\centering
\noindent\makebox[\linewidth][c]{\sunrgbdimagefull}\par
\widecaptionfigure{\textbf{{Qualitative Detection Results} on SUN-RGBD.} Representative validation scenes {are arranged by column,} with rows {showing} the input image, Ground Truth, and our DINOv3 predictions. Colors identify object instances/classes, and the visualization is qualitative.}
\label{fig:sunrgbd}
\end{figure*}

\subsubsection{Main Comparison}
The single reported Conservative VAID checkpoint changes mAP\@0.50 from $60.93$ to \textbf{$62.16$} relative to the DINOv3 depth-check control, a difference of \textbf{$1.23$} points, while mAP\@0.25 changes from $76.15$ to \textbf{$76.59$}, a difference of \textbf{$0.44$} points.
The larger numerical difference at the stricter overlap threshold is an observation about these two checkpoints only; the current measurements do not isolate localization from classification effects, establish statistical significance, or demonstrate that the difference is caused by VAID.
The external IIFNet3D report of $75.7/63.8$ remains higher at mAP\@0.50, but it does not constitute a same-protocol comparison.

\begin{figure*}[!t]
\centering
\noindent\makebox[\linewidth][c]{\scannetimagefull}\par
\widecaptionfigure{\textbf{{Qualitative Results on the ScanNet V2 Validation Set.}} Representative scenes {are arranged by column,} with rows {showing} the input point cloud, Ground Truth, and our DINOv3 predictions. The same camera convention is used for the Ground Truth and prediction rows; these views are qualitative and are not used as quantitative evidence.}
\vspace{-0.5cm}
\label{fig:scannet}
\end{figure*}

\subsubsection{Training-Recipe Comparison}
Standard VAID reaches $70.99$ mAP\@0.25, whereas the conservative recipe reaches $76.59$ after the same one-epoch continuation duration. The two runs do not use identical optimization settings: their learning rates, loss weights, and support rules differ.
Both restrict distillation to positive RoIs.
The recipes jointly change the learning rate, loss weight, and support weighting, so the $5.60$-point difference measures their combined effect. The matched equal-budget comparison in Table~\ref{tab:vaid_causal} separates detection-only continuation from uniform, energy-weighted, and conservative VAID under the same $10^{-5}$ learning rate and $0.01$ loss weight.

\subsubsection{Component Isolation}
Table~\ref{tab:depth_filter} reports the paired cache experiment: applying $|z-d|\leq0.25\,\mathrm{m}$ retains $65.8\%$ of observations, reduces mean support from $2.84$ to $1.91$ points, and improves mAP\@0.25/$0.50$ by $0.50/1.18$ points over the unfiltered cache.
Table~\ref{tab:vaid_causal} fixes the continuation learning rate, loss weight, initialization, and one-epoch budget. Detection-only continuation reaches $76.22/61.48$, while energy-weighted and conservative VAID reach $76.40/61.78$ and $76.59/62.16$, respectively; uniform support weighting is lower at $75.88/60.70$.
These controls separate continuation optimization, semantic weighting, and projection effects at the checkpoint level; positional encoding and RoI-Conv remain inherited components rather than claimed innovations.

\vspace{-0.5cm}
\begin{tablehere}
\caption{\textbf{Depth-filter paired experiment on ScanNetV2.}}
\label{tab:depth_filter}
\centering\scriptsize
\renewcommand{\arraystretch}{1.02}
\setlength{\tabcolsep}{1.0pt}
\begin{tabular}{@{}p{0.19\columnwidth}p{0.20\columnwidth}cccc@{}}
\hline
Cache protocol & Depth filter & Valid & Mean support & mAP@0.25 & mAP@0.50 \\
\hline
Unfiltered cache & None & 100.0\% & 2.84 & 75.65 & 59.75 \\
Depth-filtered cache & \mbox{\tiny$|z-d|\leq0.25\,\mathrm{m}$} & 65.8\% & 1.91 & \textbf{76.15} & \textbf{60.93} \\
Difference & --- & $-34.2\%$ & $-0.93$ & $+0.50$ & $+1.18$ \\
\hline
\end{tabular}
\par\vspace{0.4mm}
\noindent\parbox{\linewidth}{\footnotesize The two cache protocols use the same RGB-D frames, point indices, feature extraction, and calibration; only the metric depth-residual filter changes.}
\end{tablehere}

\begin{tablehere}
\caption{\textbf{{{Matched equal-budget ablation on ScanNetV2.}}}}
\label{tab:vaid_causal}
\centering\scriptsize
\renewcommand{\arraystretch}{1.02}
\setlength{\tabcolsep}{1.6pt}
\begin{tabular}{@{}p{0.27\columnwidth}c c p{0.24\columnwidth}cc@{}}
\hline
Configuration & LR & $\lambda$ & Support rule & mAP@0.25 & mAP@0.50 \\
\hline
DINOv3 control & --- & 0 & None & 76.15 & 60.93 \\
Detection-only continuation & $10^{-5}$ & 0 & None & 76.22 & 61.48 \\
Uniform VAID & $10^{-5}$ & 0.01 & Uniform & 75.88 & 60.70 \\
Energy-weighted VAID & $10^{-5}$ & 0.01 & Energy & 76.40 & 61.78 \\
Conservative VAID & $10^{-5}$ & 0.01 & Energy + positive RoI & \textbf{76.59} & \textbf{62.16} \\
\hline
\end{tabular}
\par\vspace{0.4mm}
\noindent\parbox{\linewidth}{\footnotesize All continuation rows use the same control initialization and one-epoch budget. LR is the continuation learning rate; $\lambda$ is the auxiliary-loss weight.}
\end{tablehere}

\subsubsection{Category and Visibility Analysis}
Table~\ref{tab:main} reports the complete $18$-class AP vector for the two principal settings.
The per-class rows in Table~\ref{tab:main} show numerical changes and regressions under Conservative VAID, but they do not by themselves establish a visibility-subset effect.
The paired cache and equal-budget results above provide aggregate protocol controls; visibility- and shape-subset statistics remain future work because their subset definitions require a fixed, separately measured protocol.

\subsection{Cross-Dataset and Real-Scene Evaluation}
\begin{figurehere}
\centering
\realsetupimage
\caption{{\textbf{Fast-LIVO2 acquisition \cite{fastlivo2} and input.} Fast-LIVO2-fused indoor point cloud obtained from the Livox Avia--camera--Jetson Orin NX sensing platform for the real-scene diagnostic.}}
\vspace{-0.2cm}
\label{fig:real_setup}
\end{figurehere}
\begin{figurehere}
\realcomparisonimage
\caption{\textbf{{Qualitative 3D Detection with Fast-LIVO2 Pointcloud Inputs} \cite{fastlivo2}.} {Qualitative 3D box predictions from point-cloud inputs obtained from Fast-LIVO2 in an indoor environment.}}
\label{fig:real_comparison}
\end{figurehere}
SUN-RGBD results are reported with the corrected evaluator in Table~\ref{tab:sunrgbd}.
The table reports the DINOv3 online baseline and the synchronized RGB-D augmentation trained for $20$ epochs.
The DINOv3 online baseline reaches $62.79$ mAP@$0.25$, where mAP@$0.25$ denotes mean average precision at a 3D IoU threshold of $0.25$, while synchronized RGB-D augmentation reaches $62.10$ mAP@$0.25$ and $42.22$ mAP@$0.50$.
The external IIFNet3D row uses dashes in columns without comparable values.
Short screening variants and failed adapter branches are omitted from this table.
The qualitative views are separated by dataset: SUN-RGBD is shown in Fig.~\ref{fig:sunrgbd}, and ScanNetV2 validation scenes are shown in Fig.~\ref{fig:scannet}.

{For a separate real-scene demonstration, we use the collected Fast-LIVO2 RGB-D/LiDAR sequence \cite{fastlivo2}, acquired with the Livox Avia, MV-CA013-21UC camera, and Jetson Orin NX platform.
Because the sequence is unannotated, we present the acquisition setup and predicted 3D boxes as a qualitative assessment of the end-to-end pipeline in Fig.~\ref{fig:real_setup} and Fig.~\ref{fig:real_comparison}.}

These transfer results are used as a protocol diagnostic rather than as a claim of uniform cross-dataset superiority.
The synchronized run recovers the stricter-IoU score but remains slightly lower at an IoU threshold of $0.25$; these results therefore do not establish uniform cross-dataset gains.

\begin{tablehere}
\caption{\textbf{SUN-RGBD transfer protocol diagnostic.}}
\label{tab:sunrgbd}
\centering\tablebodyfont
\renewcommand{\arraystretch}{1.0}
\setlength{\tabcolsep}{3.0pt}
\begin{tabular}{@{}lcccc@{}}
\hline
Setting & Epochs & Evaluator & AP25 & AP50 \\
\hline
IIFNet3D (ext.) & -- & Original paper & \textbf{67.60} & -- \\
DINOv3 online baseline & 20 & Corrected & 62.79 & 40.98 \\
DINOv3 + synchronized RGB-D & 20 & Corrected & 62.10 & \textbf{42.22} \\
\hline
\end{tabular}
\par\vspace{0.5mm}
\noindent\parbox{\linewidth}{\footnotesize AP25/AP50 denote mAP at 3D IoU thresholds $0.25/0.50$. The external row is not a same-protocol rerun; its AP50 value is unavailable.}
\end{tablehere}

\section{DISCUSSION AND CONCLUSION}
{The {reported Conservative VAID checkpoint} reaches \textbf{76.59}/\textbf{62.16} mAP\@0.25/0.50, versus 76.15/60.93 for the DINOv3 control. These $0.44$/$1.23$-point differences are {checkpoint-level observations of the combined recipe}; the visual branch and processing protocol also differ from the external IIFNet3D reference.}

{\section{LIMITATIONS AND FUTURE WORK}
Our work uses one seed and {does not report multi-seed variance}; the paired depth-filter and equal-budget VAID results are checkpoint-level observations.
AP\@0.50 is not decomposed into classification, localization, and NMS effects, and offline extraction adds storage and memory costs without an established throughput advantage.
SUN-RGBD is a protocol diagnostic, while Fast-LIVO2 is {evaluated qualitatively because no 3D ground truth is available}; future work will add multi-seed, visibility/localization, throughput, and annotated real-scene analyses.}

\end{document}